# Unsupervised Segmentation of Hyperspectral Remote Sensing Images with Superpixels

Mirko Paolo Barbato[a,b,c,d], Paolo Napoletano[b], Flavio Piccoli[a,b] and Raimondo Schettini[b]

[a]*Istituto Nazionale di Fisica Nucleare*
[b]*Università Milano−Bicocca*
[c]*Postal address: Viale Sarca 336, 20126, Sesto San Giovanni (MI), Italy*
[d]*Mail: m.barbato2@campus.unimib.it*

Abstract

In this paper, we propose an unsupervised method for hyperspectral remote sensing image segmentation. The method exploits the mean-shift clustering algorithm that takes as input a preliminary hyperspectral superpixels segmentation together with the spectral pixel information. The proposed method does not require the number of segmentation classes as input parameter, and it does not exploit any a-priori knowledge about the type of land-cover or land-use to be segmented (e.g. water, vegetation, building etc.). Experiments on Salinas, SalinasA, Pavia Center and Pavia University datasets are carried out. Performance are measured in terms of normalized mutual information, adjusted Rand index and F1-score. Results demonstrate the validity of the proposed method in comparison with the state of the art.

## CRediT authorship contribution statement

**Mirko Paolo Barbato:** Conceptualization, Methodology, Software, Validation, Formal analysis, Investigation, Writing − Original Draft, Writing − Review & Editing, Visualization. **Paolo Napoletano:** Conceptualization, Methodology, Software, Validation, Formal analysis, Investigation, Writing − Original Draft, Writing − Review & Editing, Visualization. **Flavio Piccoli:** Conceptualization, Methodology, Software, Validation, Formal analysis, Investigation, Writing − Original Draft, Writing − Review & Editing, Visualization. **Raimondo Schettini:** Conceptualization, Methodology, Software, Validation, Formal analysis, Investigation, Writing − Original Draft, Writing − Review & Editing, Visualization.

## 1. Introduction

The analysis of hyperspectral remote sensing images (HSI) has become more and more important in a wide number of fields, such as environmental monitoring, conservation goals, spatial planning enforcement, or ecosystem-oriented natural resources management (Blaschke (2010)). The use of HSI permits the analysis of specific electromagnetic ranges that allow a precise differentiation of observed materials, based on their spectral signatures (Murphy and Maggioni (2018)). Its ability to distinguish among several materials has shown to be a great boost in terms of performance for *HSI classification*. For example, buildings, cultivations, rivers can be easily discerned in the images as their spectral profile is different.

To fruitfully exploit HSI classification, in particular for data-hungry methods such as neural networks (Fitton et al. (2022)), we must rely on large and properly annotated image datasets. Unfortunately, labeled datasets publicly available in the state of the art are few and extremely small, mostly are composed by a single image (Yuan et al. (2021)). The main problem in the generation of remote sensing image ground truth is that the labeling is usually an interactive task that takes a lot of time and effort (Santiago et al. (2020)) and it is an operation susceptible to errors (Bahraini et al. (2021)). To overcome these problems in pixel annotations, a pre-processing of segmentation can be applied to divide the data into homogeneous regions and objects (Costa et al. (2018)). Most segmentation methods do not directly extract meaningful image objects, but clusters with generic labels, which can be used as the foundation of a succeeding procedure (Kotaridis and Lazaridou (2021)). To provide a pixelwise image segmentation that can be later exploited in an interactive labeling process to speed-up ground truth creation (Breve (2019)), different approaches of unsupervised clustering have been proposed ( Zhang et al. (2019a); Obeid et al. (2021)).

Previous studies report on the possibility of using *superpixels* to group pixel sharing homogeneous characteristics for semi-supervised image segmentation (Shen et al. (2019); Zhang et al. (2019b)). A recent and relevant segmentation method is the balanced deep embedded clustering (BDEC - Obeid et al. (2021)) that works directly on the raw hyperspectral image. This method is a variant of the deep embedded clustering (DEC - Xie et al. (2016)) adapted to work properly with imbalanced data. This method is not completely unsupervised since it requires an *a-priori* knowledge about the land to be segmented (e.g. water, vegetation, building, etc.).

In this paper, we propose an unsupervised method for hyperspectral remote sensing pixel-wise image segmentation that exploits the mean shift algorithm (Comaniciu and Meer (2002a)) that takes as input a preliminary superpixels segmentation together with the spectral pixel information. The preliminary superpixels segmentation is obtained using a modified version of the Simple Linear Iterative Clustering (SLIC) (Achanta et al. (2012)) algorithm that considers as input a concatenation between the hyperspectral image and a clustered-hyperspectral information achieved by using unsupervised clustering. The use of clustered information reduces the effect of noise, typical in hyperspectral remote sensing images.

The proposed method, differently from the state of the art, does not require the number of segmentation classes as input parameters (Wang et al. (2019)), as well as it does not require a-priori knowledge about the type of land-cover or land-use to be segmented. We demonstrate the effectiveness of the proposed method with respect to the state of the art on four publicly available datasets of hypespectral remote sensing images.

Despite its simplicity, the proposed method outperforms the state of the art in terms of average ARI and NMI. In addition, it permits to overcome some limitations found in the literature that limit the applicability in real case scenarios, in particular:

- it does not require the training of a specific neural model since it is based on handcrafted features

- it does not require a-priori knowledge about the number of classes present in the image;

- it does not require external knowledge about the image content such as the vegetation index, water etc;

Moreover, we present two variants of the proposed method, one totally automatic and one that can be easily tweaked through a single parameter to improve the performance on a new dataset.

The paper is organized as follows. In section 2 we discuss the most relevant unsupervised methods for HSI segmentation. In section 3, we describe our method starting from the general pipeline and going more in detail to each part of the pipeline. Then, in subsection 4.1, we describe the hyperspectral remote sensing datasets that we considered in our experiments. In subsections 4.2 and 4.3, we present respectively the evaluation metrics used and the results achieved by our method in comparison with the state of art. In subsection 4.4 we assess the robustness of the proposed method to several types of noise. In subsection 4.5 we present the complexity. Conclusions and future works are reported in section 5.

## 2. Related work

Methods belonging to the state of art for unsupervised and semi-supervised hyperspectral image segmentation should take into consideration both the spectral and the spatial information to avoid noisy results Liu et al. (2017); Zhang et al. (2017). Depending on the order by which these two information are addressed, it is possible to define a taxonomization of the methods composing the state of art. Table 1 shows pros and cons of each method of the state of art.

**Spatial regulation methods** Audebert et al. (2019) perform in first place a per-pixel segmentation followed by a spatial regularization done with a context-dependent criteria. In this context, Wu et al. (2016) propose a Laplacian Support Vector Machine (LapSVM) to classify each pixel and then use a Conditional Random Field (CRF) to regularize the results according to the surrounding of the pixel under consideration.

**Pre-segmentation methods** perform a first step of spatial regularization as an unsupervised pre-processing and then aggregate the spectral features for each segmented region to enforce spatial consistency. The spatial regularization is enforced through clustering or superpixels (Vargas et al. (2015)). Gillis et al. (2014) designed a fast hierarchical clustering algorithm for hyperspectral images (HNMF). The algorithm uses a rank-two nonnegative matrix factorization to split the data into clusters. This approach showed effectiveness on synthetic and real-world HSI, outperforming standard clustering techniques such as k-means, spherical k-means, and standard NMF. Zhang et al. (2019b) extend the SLIC algorithm for HSI segmentation. The authors state that the reduction of the spectrum, in their case, degrades the performance. Visual attention mechanisms can also be used to better highlight the salient parts (Haut et al. (2019); Zhang et al. (2022)).

In this context, the compression of the spectral signal with dimensionality reduction techniques such as the Principal Component Analysis (PCA), helps to decrease the size of the problem and to improve the overall performance. Zhang et al. (2017) proposed a multiscale superpixel representation starting from the first principal component. Using this representation, a multiscale classification is achieved and then fused using a majority voting to exploit the final labels. Similarly, Zhang et al. (2019a) used the Entropy Rate Superpixel segmentation and a kernel-based extreme learning machine. Starting from the first principle they successfully combined the spectral and spatial information with performance improvement over other spectral approaches. Zu et al. (2019) took advantage of the spatial information combining a band reduction technique with SLIC segmentation and a feature extraction based on principal components. This approach demonstrated to achieve results comparable to other techniques with few labeled samples.

Self-supervised methods use autoencoders to learn a compact representation of the input data (Jia et al. (2021)). To learn a meaningful representation, those methods usually use a normalization term (Coates et al. (2011)) or add noise during training (Vincent et al. (2008)). Chen et al. (2014) and Abdi et al. (2017) used stacked autoencoders to create a latent space of lower dimensionality through the use of a normalization term. Similarly, Xing et al. (2016) use stacked autoencoders but during training time they add noise to the embeddings and treat the problem as a denoising task. Nalepa et al. (2020) introduce dependency among samples through the use of 3D convolutional autoencoders. Zhang et al. (2021) use an information fusion network that combines hyperspectral images with light detection and ranging data (LiDAR). An autoencoder is trained to reconstruct both signals in a self-supervised way. Intermediate representation is then used by a two-branch CNN for final classification. Paul and Bhoumik (2022) use an U-net architecture along with spectral partitioning. The proposed architecture is called HyperUnet. Tulczyjew et al. (2020) propose the use of an asymmetric autoencoder based on recurrent neural networks to address the low cardinality and imbalance that is typical of HSI datasets. Chen et al. (2022) use adversarial training to fill the lack of samples. As stated by Wambugu et al. (2021), the generation of synthetic samples through data augmentation can improve robustness.

Graph theory is also broadly used in this context to leverage spatial relationships. In this context, Aletti et al. (2021) propose a semi-supervised method that uses a random walker method to perform segmentation. Ding et al. (2021) use a graph neural network (GNN) with autoregressive moving average filters for leveraging structures present in the HSI. Similarly, Luo et al. (2021) use GNN in combination with a multi-structure unified discriminative embedding to enforce spatial consistency.

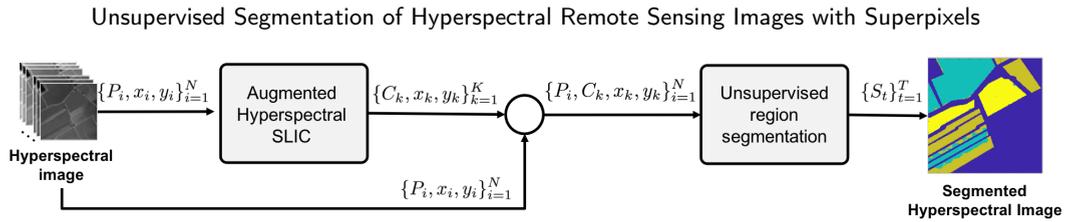

**Figure 1:** Pipeline of the proposed method. In the first step, given the hyperspectral image, the augmented hyperspectral superpixels are calculated (see Fig. 2). In the second step, centers of the superpixels along with the hyperspectral image are then used for unsupervised region segmentation using the mean shift algorithm and major voting.

**Joint learning methods** attempt to learn simultaneously spatial and spectral features. Zhang et al. (2015) developed a classification framework using gradient-fusion of bands combined with watershed superpixel segmentation to convey contextual information and spatial dependencies. By doing so, the classification will be less sensitive to noise and segmentation scales. Murphy and Maggioni (2018) proposed an unsupervised spectral-spatial diffusion learning technique (DLSS) that combines spectral and spatial information considering the modes of classes. This active learning strategy can be helpful in those contexts where the hyperspectral information varies along time and the system must adapt to new data. Xie et al. (2016) proposed a Deep Embedded Clustering (DEC) to simultaneously learn feature representations and cluster assignments using deep neural networks. DEC maps the data space to a lower-dimensional feature space optimizing a clustering objective. The experimental evaluations on state of art images showed significant improvement. Starting from this approach, Obeid et al. (2021) proposed a balanced version of DEC (BDEC). In particular, they developed an additional search and extraction step to balance the data before the training of DEC, making use of a-priori knowledge on the context of data and labeling, to further improve the overall quality of the segmentation on a variety of state-of-the-art datasets. They compared their method with other clustering techniques such as k-means (Obeid et al. (2021)), Gaussian mixture model (GMM, Obeid et al. (2021)), and sparse manifold clustering and embedding (SMCE, Elhamifar and Vidal (2013)).

## 3. Proposed method

Figure 1 shows the pipeline of the proposed method. It can be divided into two steps: 1) augmented hyperspectral superpixels are achieved by using a modified version of the Simple Linear Iterative Clustering (SLIC) that takes as input the hyperspectral image and a clustered-hyperspectral information achieved by using unsupervised clustering ; 2) augmented superpixels along with the hyperspectral image are used by an unsupervised region segmentation module to achieve the final segmentation. In the following, each step of the pipeline is presented.

### 3.1. Augmented Hyperspectral SLIC superpixels

Hyperspectral remote sensing images are usually very large images that, due to the imaging conditions, may be afflicted by different artifacts. As a consequence, superpixels calculated on hyperspectral images may present imprecise boundaries. To improve the goodness of superpixels, we propose a modification of the original SLIC algorithm that is composed of two steps: a preliminary unsupervised clustering using the mean-shift algorithm (Finkston (2022)) whose output (that we call clustered-hyperspectral information) is used together with the original image by the SLIC algorithm.

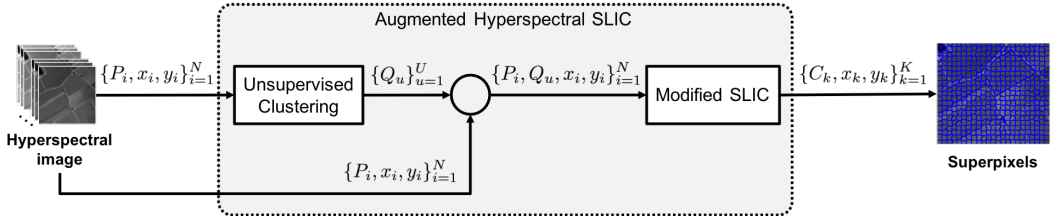

**Figure 2:** Augmented Hyperspectal Simple Linear Iterative Clustering (SLIC). In the first step, given the hyperspectral image, an unsupervised clustering is applied and the output is merged with the input hyperspectral image to finally achieve the augmented hyperspectral superpixels.

### 3.1.1. Unsupervised clustering

The pipeline of the augmented superpixels method is depicted in Figure 2. Given an input hyperspectral image I made of $N$ pixels $P_i = (b_1^i, \cdots, b_L^i)$ and $L$ bands, we extract pixel similarity information using the mean-shift algorithm (Comaniciu and Meer (2002b)) which is an unsupervised clustering algorithm that iteratively finds the best number of clusters $U$ that better fits the input data. Each cluster center is defined by the mean of pixels of the hyperspectral image that are assigned to the $u$-th cluster defined as follows: $Q_u = (q_1^u, \cdots, q_L^u)$.

To avoid outliers in the spectral signal, a normalization of the hyperspectral image is performed by considering its maximum value $V$ as for the 95% of the spectral data. All the values of the image are clipped between 0 and $V$ and then divided by the same value $V$ so that the final image is normalized between 0 and 1.

Once the algorithm has found the clusters, the spectral information of each pixel of the image is concatenated with the center of the cluster to which it belongs to. The resulting vector describing a pixel at a position $x_i$ and $y_i$ is therefore $F_i = \langle P_i, Q_u, x_i, y_i \rangle$ which is of a size equal to $(2 \times L) + 2$. The augmented hyperspectral image $\{P_i, Q_u, x_i, y_i\}_{i=1}^N$ is the input of a modified version of the SLIC method.

### 3.1.2. Modified SLIC

The original SLIC algorithm takes as input RGB color and spatial information of the image, namely $N$ pixels $P_i = (b_R^i, b_G^i, b_B^i)$ at position $x_i$ and $y_i$, and it exploits the $k$-means algorithm to cluster them into superpixels (Achanta et al., 2012). The algorithm initially considers a number $K$ of superpixel cluster centers $C_k$ taken at regular grid intervals $S = \sqrt{N/K}$. The higher is the number of $K$ and the smaller is the size of the initial superpixels. Ideally, $S^2$ represents the area of each superpixel. To assign the pixel to the cluster $k$ a search region of $2S \times 2S$ around the cluster center is used. This strategy reduces the complexity of the SLIC algorithm so that it is linear to the number of pixels $N$ and independent from the number of superpixels $K$.

After the initialization of the grid, the algorithm iteratively assigns each pixel $P_i$ to the nearest superpixel, whose search area overlaps the pixel itself. For each iteration, the assignment is determined by a distance that, in the original formulation of the SLIC algorithm, is defined as follows:

$$D_s = d_{rgb} + m\left(\frac{d_{xy}}{S}\right) \tag{1}$$

where $m$ is a parameter used to control the regularity of superpixels using spatial information, while $d_{rgb}$ and $d_{xy}$ are respectively, the color and spatial differences between a pixel and the center of the corresponding superpixel.

The algorithm repeats the assignment process between pixels and superpixels until the Euclidian distance between the old centers and the new centers is lower than a certain threshold.

---

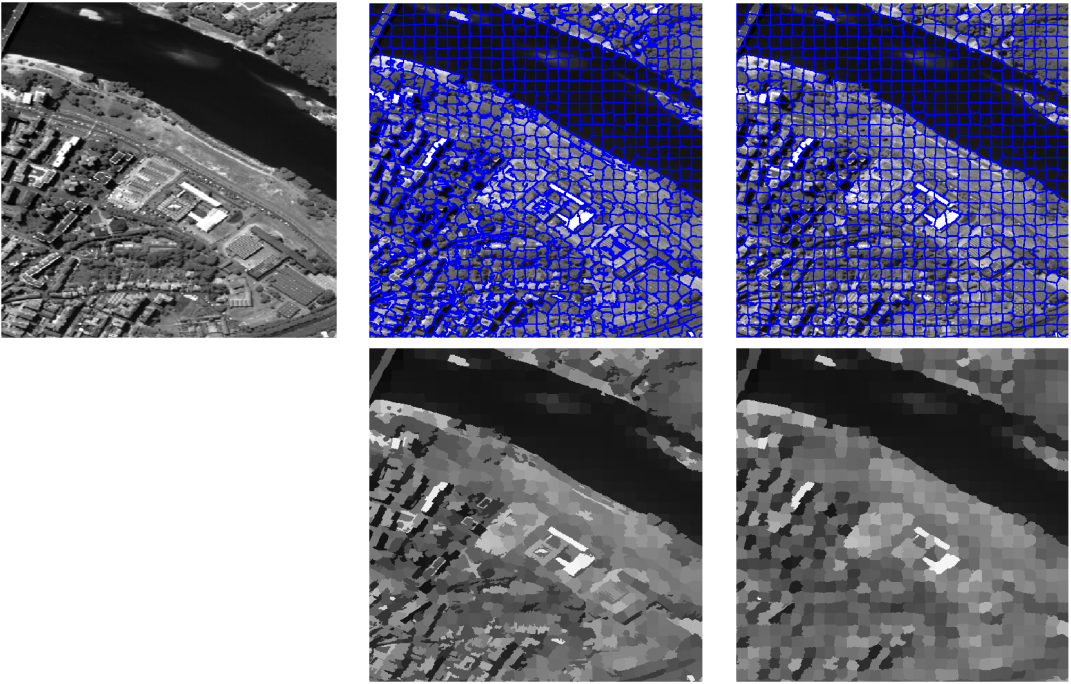

**Figure 3:** On the left it is possible to see a projection of a portion of the Pavia Center dataset on a single band, while in the center and on the right it is shown the results of the modified SLIC algorithm, respectively with $m = 0.2$ and $m = 1$. In these examples we use a fixed valued for $m_{clust} = 0$

In our modified version of the SLIC algorithm, instead of $D_s$, we define a new distance $D_{ahs}$ in order to handle the hyperspectral image and the clustered hyperspectral information at the same time. The new distance $D_{ahs}$ between a pixel $F_i$ and $k$-th superpixel center $C_k$ is defined as follows:

$$D_{ahs} = \frac{d_{spec}}{\sqrt{L}} + m_{clust}\left(\frac{d_{clust}}{\sqrt{L}}\right) + m\left(\frac{d_{xy}}{S\sqrt{2}}\right) \tag{2}$$

where

$$d_{spec} = \sqrt{\sum_{j=1}^{L}(b_j^k - b_j^i)^2} \tag{3}$$

$$d_{clust} = \sqrt{\sum_{j=1}^{L}(q_j^k - q_j^u)^2} \tag{4}$$

and

$$d_{xy} = \sqrt{(x_k - x_i)^2 + (y_k - y_i)^2} \tag{5}$$

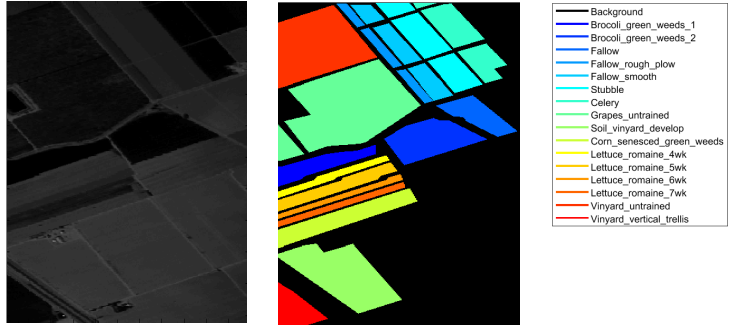

**Figure 4:** On the left it is possible to see a projection of the Salinas dataset on a single band, while on the right its ground truth and the corresponding color labeling. Note that the black color represents the background.

To make the distance $D_{ahs}$ independent from the number of bands, we have normalized the distances $d_{spec}$ and $d_{clust}$ by a factor $\sqrt{L}$ that is achieved considering the maximum L2 distance between two most diverse hyperspectral pixels. Since each hyperspectral band ranges between 0 and 1, the most diverse pixels are the following: $P_0 = (0, \cdots, 0)$ and $P_1 = (1, \cdots, 1)$. The $d_{spec}$ (as well as $d_{clust}$) between these two pixels is therefore $\sqrt{L}$.

The same idea is applied for the normalization of the spatial information. Considering the standard search region of the SLIC algorithm, the maximum spatial distance between a superpixel center and a pixel in the search region is $2S/\sqrt{2} = S\sqrt{2}$.

The parameters $m$ and $m_{clust}$ are used to control the regularity of superpixels using spatial and clustered hyperspectral information respectively. Figure 3 shows an example of the proposed SLIC achieved on a hyperspectral image as $m$ varies between 0 and 1. As you can see the higher is the value of $m$, the higher is the compactness of the superpixels that tend to resemble the square shape of standard pixels. In these examples we use a fixed valued for $m_{clust} = 0$, but analogous considerations can be done by varying the parameter $m_{clust}$.

### 3.2. Unsupervised region segmentation

This module takes as input the concatenation of the original hyperspectral image $\{P_i = (b_1^i, \cdots, b_L^i)\}_{i=1}^N$ with the corresponding centers of our superpixels $\{C_k = (b_1^k, \cdots, b_L^k, x_k, y_k)\}_{k=1}^K$ that are the average color (over the $L$ bands of the hyperspectral image) of the pixels and their spatial positions. The input, in short $\langle P_i, C_k, x_k, y_k \rangle$, is processed by the mean-shift algorithm (Finkston (2022)) along with a major voting strategy to generate the final segmentation map $\{S_t\}_{t=1}^T$.

The use of mean-shift is motivated by the fact that, differently from other clustering methods, the knowledge of the number of clusters to be predicted is not required, moreover it demonstrated to reduce the number of mislabeled samples with respect to other methods in the state of the art (Bahraini et al., 2021).

At the end of the clustering, we further improved the segmentation results by eliminating small regions that are likely due to noise. This is done by re-assigning a label to a given region that is more frequent in its neighborhood.

## 4. Experimental validation

### 4.1. Datasets

The first two datasets used are Salinas and SalinasA (respectively in figure 4 and 5) (M Graña (2020)). These images have been collected by the AVIRIS sensor and present 224 bands in the 400–2500nm portion of the spectrum, where 20 noisy bands in the region of water absorption have discarded, resulting in two images with 204 channels. The dimensions of the datasets are respectively 512x217 and 86x83 with a high spatial resolution of 3.7m for pixels. The labeling is different between the two datasets. Salinas presents 16 classes that represent the region of culture such as broccoli, fallow, grapes, etc., while SalinasA is segmented in a subset of the former dataset, representing only 6 classes from broccoli to corn and different variations of lettuce.

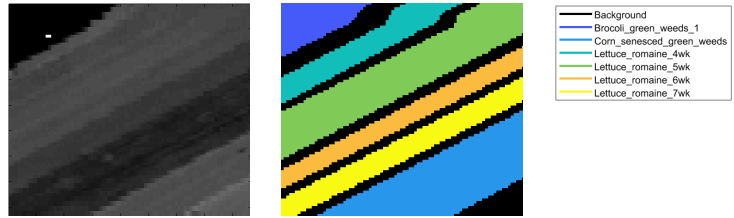

**Figure 5:** On the left it is possible to see a projection of the SalinasA dataset on a single band, while on the right its ground truth and the corresponding color labeling. Note that the black color represents the background.

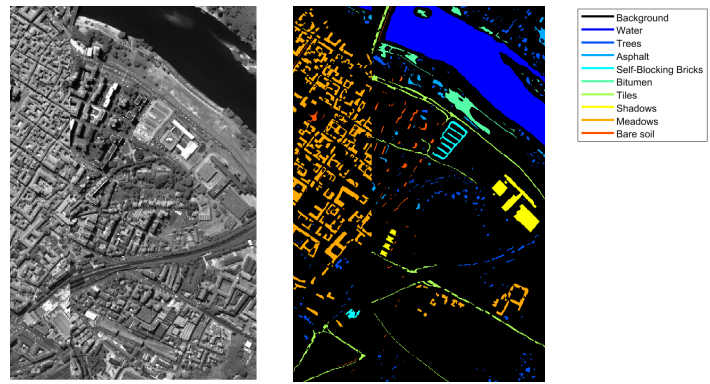

**Figure 6:** On the left it is possible to see a projection of the Pavia Center dataset on a single band, while on the right its ground truth and the corresponding color labeling. Note that the black color represents the background.

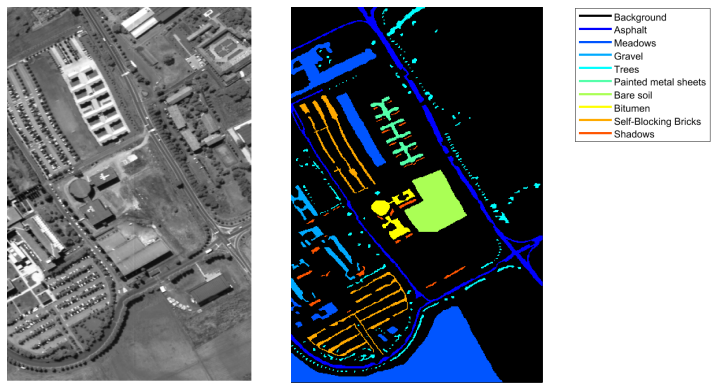

**Figure 7:** On the left it is possible to see a projection of the Pavia University dataset on a single band, while on the right its ground truth and the corresponding color labeling. Note that the black color represents the background.

We also considered on the Pavia Center and Pavia University datasets (M Graña (2020)) that respectively are images of 1096x1096 with 102 channels and 610x610 with 103 channels (see figure 6 and 7). However, in both images part of the samples have been discarded because the information was missing, resulting in two images respectively of 1096x715 and 610x340. The two scenes have been acquired by the ROSIS sensor with a geometric resolution of 1.3 meters. These datasets are both annotated with 9 labels typical of a city such as asphalt, meadows, trees, bare soil, etc..

## 4.2. Evaluation metrics

Our method is composed by two modules. The former segments the input image using superpixels while the latter clusters data using spectral and spatial information to perform the final segmentation. Following the guidelines of Achanta et al. (2012), we evaluated the superpixel segmentation step with the UE (Undersegmentation Error), while the second step has been evaluated by using ARI (Adjusted Rand Index), NMI (Normalized Mutual Information) and F-measure (Obeid et al. (2021)).

### 4.2.1. Superpixel Segmentation - Undersegmentation error

An undersegmentation error (UE) occurs when pixels belonging to different classes considered in the task are grouped together into a single region/class. Given a region of the ground truth $g_i$, the UE is defined as follows:

$$UE = \frac{1}{N} \left[ \sum_{i=1}^{M} \left( \sum_{S_j | S_j \cap g_i > B} |S_j| \right) - N \right] \tag{6}$$

where $M$ is the number of ground truth segments, $B$ is a minimum number of pixels in $S_j$ overlapping $g_i$ and $N$ is the number of pixels of the image. $B$ is used to compensate for possible errors in the ground truth segmentation data. The lower is the UE and the better is the method.

### 4.2.2. Unsupervised region segmentation

To measure the performance of the whole method, three evaluation metrics have been used: normalized mutual information (NMI), adjusted Rand index (ARI) (Pedregosa et al. (2011); Sheng and Huber (2020)), and F1-score (Mallawaarachchi et al. (2020)). NMI and ARI are defined as:

$$NMI = \frac{\sum_i \sum_j n_{ij} \log(\frac{n \cdot n_{ij}}{n_i \cdot n_j})}{\sqrt{\sum_i n_i \log \frac{n_i}{n} \sum_j n_j \log \frac{n_j}{n}}} \tag{7}$$

$$ARI = \frac{\sum_{ij} \binom{n_{ij}}{2} - [\sum_i \binom{n_i}{2} \sum_j \binom{n_j}{2}]/\binom{n}{2}}{\frac{1}{2}[\sum_i \binom{n_i}{2} + \sum_j \binom{n_j}{2}] - [\sum_i \binom{n_i}{2} \sum_j \binom{n_j}{2}]/\binom{n}{2}} \tag{8}$$

where $n$ is the total number of samples, $n_i$ is the number of samples in a cluster $i$, $n_j$ is the number of samples in class $j$, and $n_{ij}$ is the number of samples in both clusters $i$ and class $j$. For both the NMI and ARI, the higher is the value the better is the method.

F1-score ( Mallawaarachchi et al. (2020)) for unsupervised clustering, where the number of classes is unknown, is defined as:

$$Precision = \frac{\sum_k \max_s \{a_{ks}\}}{\sum_k \sum_s a_{ks}} \tag{9}$$

$$Recall = \frac{\sum_s \max_k \{a_{ks}\}}{\sum_k \sum_s a_{ks}} \qquad (10)$$

$$F1 = 2 \times \frac{Precision \times Recall}{Precision + Recall} \qquad (11)$$

where $k$ is the number of clusters predicted, $s$ is the number of classes, and $a_{ks}$ denotes the number of samples clustered to cluster $k$ and belonging to class $s$.

## 4.3. Results

Results section is divided into two subsections, the first one discusses the results achieved with just our augmented version of the superpixel segmentation while the second one presents the results with our entire pipeline applied for unsupervised region segmentation.

### 4.3.1. Superpixel segmentation

The metric adopted for the evaluation of the superpixel segmentation is the under-segmentation error (UE), which strongly depends on the precision of both the segmentation and the ground-truth on the boundaries (Achanta et al., 2012). Since of the most precise ground-truth annotation is available for the Salinas and SalinasA datasets, we focus here only on these datasets. Due to the imprecision of the ground-truths on the boundaries derived from the remote sensing nature of the images, we fixed the parameter $B$ of the equation (6) to the 15% of the pixels in $S_j$.

In our experiments, we set the number of superpixels required to $K = 1000$ for Salinas, while $K = 500$ for SalinasA. The bandwidth for the clustering has been empirically fixed to 0.1 for both datasets.

Our augmented SLIC segmentation has two weight parameters: spatial $m$ and clustered information $m_{clust}$. By setting the value $m_{clust} = 0$ our augmented SLIC is turned into the original SLIC applied to an hyperspectral image.

In the table 2 (a), we present the UE achieved by our algorithm with different values of $m$ and $m_{clust}$ on Salinas A. The lower is the value and the better is the result. The first raw corresponds to the original SLIC while the other raws corresponds to our algorithm with different settings of the parameter $m_{clust}$. Whatever is the value of $m$, with $m_{clust}$ higher than zero, we always achieve better results than the original SLIC. As it is possible to observe in the table, the best value is when $m = 0.2$ and $m_{clust} = 0.8$.

The table 2 (b) reports the UE on Salinas and it also demonstrates and confirms that considering also the clustered information improves the results of the superpixel segmentation. Even in this case, we achieved the best overall results when $m = 0.2$ and $m_{clust} = 0.8$.

### 4.3.2. Unsupervised region segmentation

We have evaluated our entire pipeline for unsupervised segmentation on all the four datasets described in section 4.1. For the sake of comparison, we have evaluated the unsupervised region segmentation by considering different types of input information (see pipeline in Figure 1) using both the k-means (assuming known the number of classes) and the mean-shift:

- hyperspectral information: the clustering algorithm applied directly to the input hyperspectral image;

- superpixels centers: the clustering algorithm applied to the superpixels $C_k$;

- hyperspectral and superpixels centers: the clustering algorithm applied to the concatenation of superpixels $C_k$ and the input hyperspectral image;

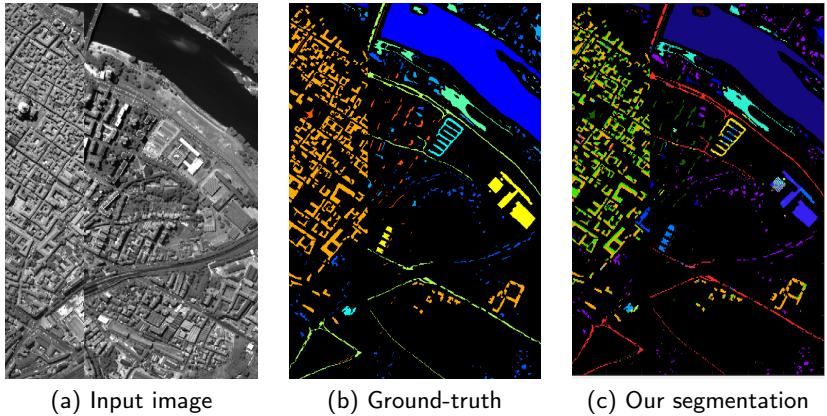

(a) Input image        (b) Ground-truth        (c) Our segmentation

**Figure 8:** On the left there is the Pavia Center dataset, in the center it is shown the Pavia Center ground truth, and on the right the unsupervised region segmentation results with the best NMI achieved. The perfomance on the metrics are $ARI = 0.88$, $NMI = 0.87$ and $F1 = 0.90$.

- reduced hyperspectral information: the clustering algorithm applied to the output of the feature reduction phase that is based on Principal Component Analysis (PCA) applied to the input hyperspectral image;

- reduced superpixels centers: the clustering algorithm applied to the superpixels centers achieved on the reduced hyperspectral information;

- reduced hyperspectral and superpixels centers: the clustering algorithm applied to the concatenation of superpixels $C_k$ and the input hyperspectral image after a feature reduction using PCA;

When using PCA we consider the 99.9% of the total variance represented by each principal component. In the table 3, it is shown the reduction of the number of bands after the application of the PCA for each dataset.

In the tables 4, 5, 6, and 7, we compare mean-shift methods with k-means applied with the correct and previously known number of clusters. All the experiments have the same values fixed for the parameters $m = 0.4$ and $m_{clust} = 0.8$ but a different number of superpixels dependant from the dataset (2000 for Pavia center and Pavia University, 800 for Salinas, 300 for SalinasA). We empirically choose the bandwidth parameter of mean-shift with the better NMI performance. The figures 8, 9, 10, 11 show the results of the proposed algorithm, considering that the color map of the labels between ground-truth and segmentation is not the same.

The results show that for each of the datasets we achieve better results when the combination of mean-shift, spectral information, and superpixel centers is involved. PCA allows the reduction in dimensionality of the feature vectors, but has not shown coherent improvements on all the datasets.

Results reported above have been obtained by heuristically tuning the following parameters:

- the bandwidth of mean-shift

- the weight $m$ of the spatial information

- the weight $m_{clust}$ of the spectral similarity information

The bandwidth of mean-shift controls the number of classes that are extracted by the clustering algorithm. The higher is the bandwidth, the lower the number of clusters.

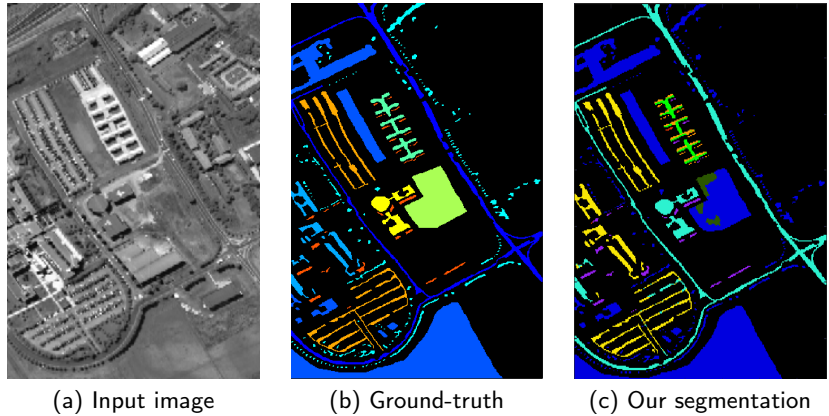

(a) Input image      (b) Ground-truth      (c) Our segmentation

**Figure 9:** On the left there is the Pavia University dataset, in the center it is shown the Pavia University ground truth, and on the right the unsupervised region segmentation results with the best NMI achieved. The perfomance on the metrics are $ARI = 0.59$, $NMI = 0.72$ and $F1 = 0.84$.

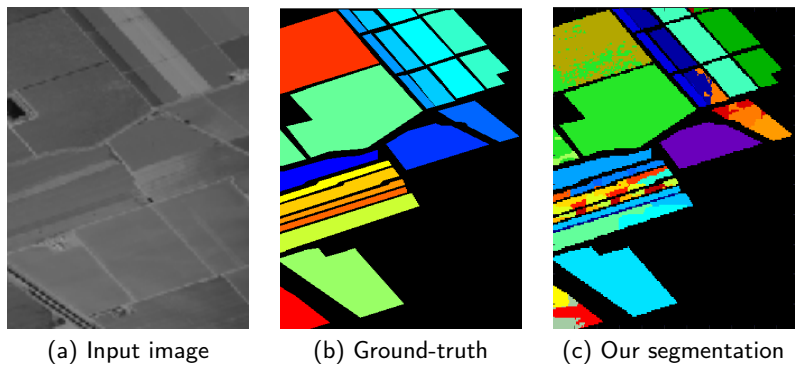

(a) Input image      (b) Ground-truth      (c) Our segmentation

**Figure 10:** On the left there is the Salinas dataset, in the center it is shown the Salinas ground truth, and on the right the unsupervised region segmentation results with the best NMI achieved. The perfomance on the metrics are $ARI = 0.85$, $NMI = 0.91$ and $F1 = 0.90$.

The parameter $m$ modulates the compactness of superpixels, so lower values are more suitable for high resolution images, while $m_{clust}$ is related to smoothing and therefore to noise reduction.

The number $K$ of superpixels depends mostly on the dimensions of the hyperspectral image, with $K$ that increases in correspondence with larger and cluttered images. We empirically found that the optimal value for $K$ is given by the approximation to the nearest hundred of the ratio between the smallest image dimension and a scaling parameter $c_1$:

$$K = \lceil \frac{min(H, W)}{\alpha \cdot 100} \rceil \cdot 100 \qquad (12)$$

where $H$ is the height, $W$ is the with of the image and $\alpha$ is the scaling factor that values 60. $K$ ranges from a minimum of 300 and a maximum of 2000 superpixels.

We also experimented with a variant of our method that exploits bandwidth values automatically determined by following the procedure proposed by Pedregosa et al. (2011). We refer to this version of the proposed method as `OUR_BW`, while we refer to the version of the proposed method with all the parameters heuristically set as `OUR`.

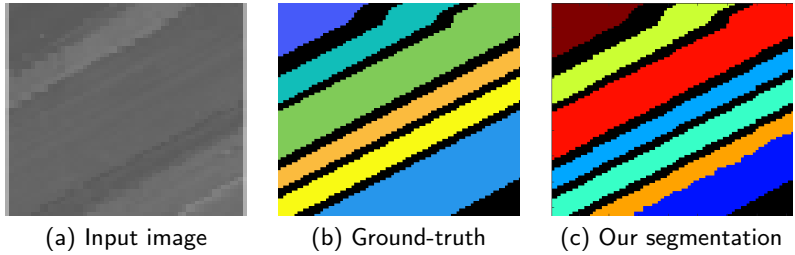

| (a) Input image | (b) Ground-truth | (c) Our segmentation |

**Figure 11:** On the left there is the SalinasA dataset, in the center it is shown the SalinasA ground truth, and on the right the unsupervised region segmentation results with the best NMI achieved. The perfomance on the metrics are $ARI = 0.90$, $NMI = 0.95$ and $F1 = 0.95$.

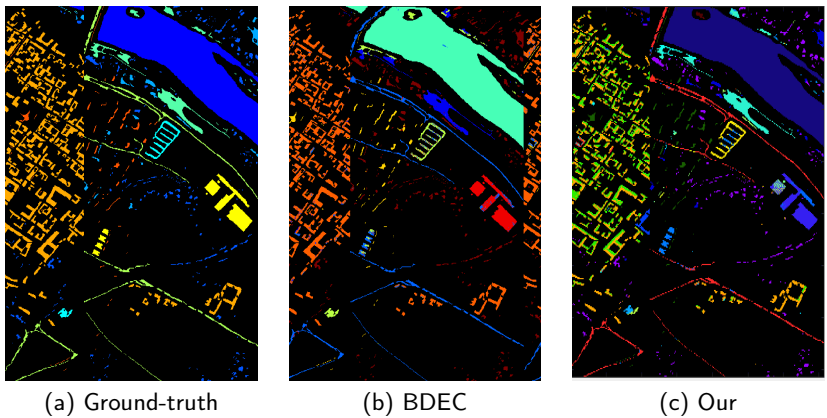

| (a) Ground-truth | (b) BDEC | (c) Our |

**Figure 12:** On the left there is the Pavia Center ground truth, in the center the results of the BDEC method, and on the right the results achieved by our technique. The perfomance of BDEC are $ARI = 0.97$ and $NMI = 0.91$. The perfomance of Our method are $ARI = 0.88$ and $NMI = 0.87$.

Finally, in table 8 we compare the ARI and NMI performance of the two variants of our method with other unsupervised segmentation methods in the state of the art (Obeid et al., 2021). All the state-of-the-art methods, whose results are reported in table 8, require the tuning of some parameters, some form of a-priori knowledge or a training process.

Our method does not need to know the exact number of classes to be identified and does not use any external knowledge about the image content and, more importantly, it is an handcrafted method that does not require the training of a specific model.

Our proposal achieves, on average, the best results in terms of NMI and ARI on the considered datasets. In particular, it outperforms all the other methods on Salinas, SalinasA, and Pavia University, while it achieves lower performance on Pavia Center. The algorithm version with automatic bandwidth achieves on average comparable results with respect to other methods.

In figures 12, 13, 14, and 15 it is shown a comparison between the segmentation results achieved by our method and BDEC technique (Obeid et al. (2021)). The results of BDEC have been retrieved by using the code available from the respective article and by reconstructing the segmented images.

### 4.4. Robustness to Noise

Taking inspiration from Nalepa et al. (2021), we assessed the robustness of our method to several types of noise. For each dataset, we have defined a new hyperspectral image $I'$ which is the result of adding noise $N$ to the hyperspectral images $I$ as follows: $I' = I + N$.

The considered types of noise are:

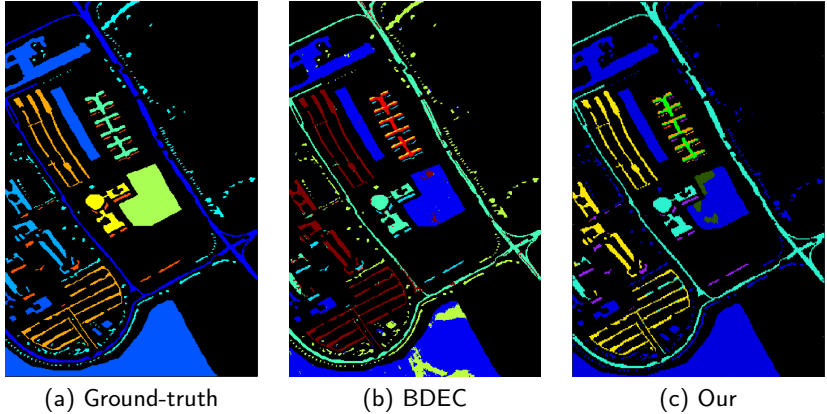

(a) Ground-truth    (b) BDEC    (c) Our

**Figure 13:** On the left there is the Pavia University ground truth, in the center the results of the BDEC method, and on the right the results achieved by our technique. The perfomance of BDEC are $ARI = 0.60$ and $NMI = 0.70$. The perfomance of Our method are $ARI = 0.59$ and $NMI = 0.72$.

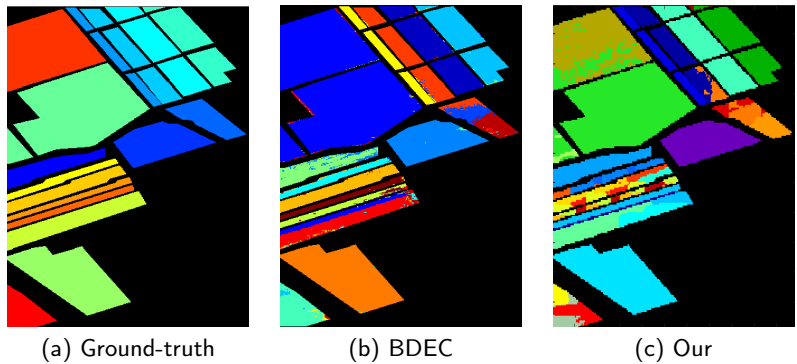

(a) Ground-truth    (b) BDEC    (c) Our

**Figure 14:** On the left there is the Salinas ground truth, in the center the results of the BDEC method, and on the right the results achieved by our technique. The perfomance of BDEC are $ARI = 0.68$ and $NMI = 0.87$. The perfomance of Our method are $ARI = 0.85$ and $NMI = 0.91$.

- Gaussian noise

- Impulsive noise (salt & pepper)

- Poisson noise

#### 4.4.1. Gaussian noise

The Gaussian noise simulates thermal and quantization disturbances (Nalepa et al., 2021). The noise signal is defined by a normal distribution probability density function. The probability $p$ for a variable $x$, with mean $\mu$ and the variance $\sigma$ is defined as:

$$p(x) = \frac{1}{\sigma\sqrt{2\pi}}e^{\frac{-(x-\mu)^2}{2\sigma^2}} \tag{13}$$

In our experiments, we have applied noise to the 10% of the pixels of the image and, for each of the evaluations, we have considered the mean $\mu = 0$ and different values of variance $\sigma$: 0, 0.01, 0.05, 0.1, 0.25, and 0.5. We show the results of this investigation in table 9.

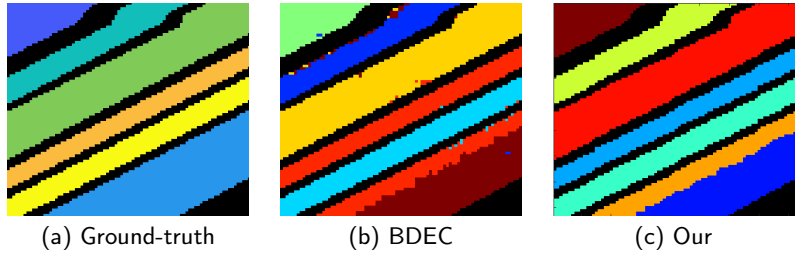

| (a) Ground-truth | (b) BDEC | (c) Our |

**Figure 15:** On the left there is the Pavia SalinasA ground truth, in the center the results of the BDEC method, and on the right the results achieved by our technique. he perfomance of BDEC are $ARI = 0.81$ and $NMI = 0.87$. The perfomance of Our method are $ARI = 0.90$ and $NMI = 0.95$.

As expected, the results show a reduction in performance when the noise is particularly disturbing. This is particularly true in the case of Salinas and SalinasA datasets when a high variance is considered. Overall, the results show that our method is robust to Gaussian noise signal when the original signal is not too deteriorated.

### 4.4.2. Impulsive noise

The impulsive noise represents an error in the acquisition of data, where a pixel remained saturated ("white" pixel) or where the data for a certain pixel is lost ("black" pixels) (Nalepa et al., 2021).

In our experiments, we have applied salt & pepper noise with density on the image. Specifically, we have considered 0, 0.01, 0.05, 0.1, 0.25, and 0.5 as the percentage of the pixels affected by noise. In the table 10, we show the results for every density and dataset.

Results show that this kind of noise does not impact significantly on the overall performance. Oscillatory performance is justified by the stochasticity of the entire process.

### 4.4.3. Poisson noise

The Poisson noise models a signal-dependent photon noise (Nalepa et al., 2021). The signal is defined as the following probability density function $p$ for a variable $x$:

$$p(x) = \frac{e^\lambda \lambda^x}{x!} \tag{14}$$

where $\lambda$ represents the expected average value, which we considered to be 5.5 for all the experiments.

In the table 11, we show the results for every dataset with and without noise.

The results show that our method is robust to the presence of Poisson noise on each dataset.

## 4.5. Computational time and complexity

In this section we show the real time computation for all of the datasets, and the complexity of the entire method. The time complexity of the whole method is $O(n^2)$ where $n$ is the number of pixels of the image. In the table 12, it is shown the time in seconds for each part of our method: Augmented H-SLIC, Unsupervised Segmentation, and the Total algorithm. Note that, on average, the time required by the augmented H-SLIC is roughly 57% of the total amount of time required for running the algorithm. Note also that the computational time depends on the size of the image, for example the execution of the algorithm on the SalinasA dataset requires about $63s$ which is eleven times lower than the Salinas dataset.

## 5. Conclusion

With the increase of available data from earth observation drones and satellites, it is very important to reduce the effort on segmenting/labeling remote sensing images. To this end we need methods that are not based on large training set, such as Deep Learning methods, that do not require a-priori knowledge and/or the number of classes to be segmented. In this paper, we have presented a method based on hand-crafted features that satisfy the above requirements. The method has been experimented on four different datasets thus proving its effectiveness in comparison with methods in the state of the art that in contrast they may not satisfy all the above requirements.

A further feature of the proposed method is that it could be easily extended to deal with additional information obtained by other types of sensors. It is also worth to be investigated if a-priori knowledge about the image content can be exploited to improve the results, in particular in urban scenes.

## Acknowledgement

Research developed in the context of the project PIGNOLETTO − Call HUB Ricerca e Innovazione CUP (Unique Project Code) n. E41B20000050007, co-funded by POR FESR 2014-2020 (Programma Operativo Regionale, Fondo Europeo di Sviluppo Regionale – Regional Operational Programme, European Regional Development Fund).

## Computer Code Availability

The source code to achieve segmentation and evaluation is available at `https://github.com/mpBarbato/Unsupervised-Segmentation-of-Hyperspectral-Remote-Sensing-Images-with-Superpixels`.
Developer Contacts:
   Name: Mirko Paolo Barbato
   Mail: m.barbato2@campus.unimib.it
   Postal address: Viale Sarca 336, 20126, Sesto San Giovanni (MI), Italy
   Requirements and code information
   Name: Unsupervised-Segmentation-of-Hyperspectral-Remote-Sensing-Images-with-Superpixels
   Year: 2022
   Hardware required: Intel(R) Core(TM) i7-6500U CPU @ 2.50GHz-2.59 GHz, RAM 8.00GB
   Software required: Windows 10 64-bit, Matlab R2021A, Python 3.7, scikit-learn
   Program language: MATLAB, Python
   Program size: 35,8 KB

Unsupervised Segmentation of Hyperspectral Remote Sensing Images with Superpixels

| Method | Pros | Cons |
|---|---|---|
| Abdi et al. (2017) | The architecture is very simple and easily replicable | The use of NNs instead of CNNs makes the system more data-hungry |
| Aletti et al. (2021) | Consensus-based methods show more robustness to noise | Dataset-tailored similarity index can lead to misjudgement |
| Chen et al. (2014) | Self-supervised approach helps to deal with small amount of data | The size of the neighbor region has a huge impact on the performance of the system and it is dataset-dependent |
| Chen et al. (2022) | The local manifold learning helps to discover relationships among samples | Adversarial prediction can face a partial or a total mode collapse |
| Ding et al. (2021) | Graph neural networks can be effective in describing structures | Requires a great amount of labeled data that often is not available |
| Elhamifar and Vidal (2013) | Through the use of sparse coding the method is robust to data nuisances such as noise and outliers | the refactorization of each sample as a linear combination of remaining samples can be misleading in presence of many outliers |
| Gillis et al. (2014) | Nonnegative matrix factorization can be a powerful splitting technique | Requires to know in advance the number of classes |
| Liu et al. (2017) | Only five parameters required and a moderate number of training samples | High computational time both in training and testing due to the construction of the similarity graph |
| Luo et al. (2021) | Intraclass and interclass neighborhood structure graphs can help to improve the description in the feature space of the HSI | It's unclear how much is the contribution of the tangential structure information on the final performance |
| Murphy and Maggioni (2018) | The proposed algorithm can be complemented with few real samples, boosting the performance in an active learning fashion | Based on the assumption that different classes have different densities |
| Nalepa et al. (2020) | Representation learned in a self-supervised fashion | 3D convolutions introduce in-batch samples dependency |
| Obeid et al. (2021) | Extremely fast (up to 2600X w.r.t. Xie et al. (2016)). Address the problem of data imbalance | The extraction of data subsets that are equally representative can lead to misrepresentation in presence of low cardinality classes |
| Vargas et al. (2015) | Fast and simple as the computation is limited to neighbors | Generalization limited by bag of words which must be small to avoid a performance drop |
| Wu et al. (2016) | Pixelwise classification is simple and fast | low resolution of hyperspectral images leads to pixel misclassification |
| Xie et al. (2016) | Feature representation is learned during the process. Less sensitive to hyperparameters | Kullback-Leibler divergence minimization can lead to errors when the auxiliary distribution has low cardinality |
| Xing et al. (2016) | Very simple architecture | Poor results |
| Zhang et al. (2015) | The proposed fusion method requires less samples with respect to other methods | The use of Local Binary Patterns (LBPs) heavily affects the performance as the scale changes. This is due to the limited field of view of LBPs. |
| Zhang et al. (2017) | The multiscale approach avoids the choice of the optimal superpixel size | Major voting strategy considers all scales equally |
| Zhang et al. (2019b) | Exploits the consolidated SLIC algorithm to define superpixels, extending it to hyperspectral information | It is semi-supervised and requires some labeled samples to propagate their label to the pixels in the same superpixel |
| Zhang et al. (2019a) | Captures local as well as global spatial characteristics of the HSI | Extreme learning machine (EML) uses single hidden layer feedforward neural networks (SLFNs), whose representativity is low |
| Zhang et al. (2021) | Improves performance by integrating LiDAR data | requires LiDAR data |
| Zu et al. (2019) | Lower dimensionality promotes meaningfulness of feature vectors | Superpixel-independent dimensionality reduction through robust PCA can lead to non-comparable feature vectors |

**Table 1**
Pros and cons of the methods representing the state of the art.

|  | $m = 0.2$ | $m = 0.4$ | $m = 0.6$ | $m = 0.8$ | $m = 1$ |
|---|---|---|---|---|---|
| $m_{clust} = 0$ | 0.2148 | 0.2102 | 0.2102 | 0.2098 | 0.2098 |
| $m_{clust} = 0.2$ | 0.2088 | 0.2073 | 0.2067 | 0.2067 | 0.2067 |
| $m_{clust} = 0.4$ | 0.2051 | 0.2080 | 0.2071 | 0.2067 | 0.2067 |
| $m_{clust} = 0.6$ | 0.2053 | 0.2088 | 0.2073 | 0.2067 | 0.2067 |
| $m_{clust} = 0.8$ | **<u>0.2030</u>** | 0.2061 | 0.2047 | 0.2043 | 0.2040 |
| $m_{clust} = 1$ | 0.2055 | **0.2050** | **0.2032** | **0.2024** | **0.2018** |

(a) SalinasA

|  | $m = 0.2$ | $m = 0.4$ | $m = 0.6$ | $m = 0.8$ | $m = 1$ |
|---|---|---|---|---|---|
| $m_{clust} = 0$ | 0.2766 | 0.2856 | 0.2938 | 0.2998 | 0.3019 |
| $m_{clust} = 0.2$ | 0.2714 | 0.2845 | 0.2986 | 0.3087 | 0.3066 |
| $m_{clust} = 0.4$ | 0.2661 | 0.2867 | 0.2954 | 0.3011 | 0.3065 |
| $m_{clust} = 0.6$ | 0.2649 | 0.2828 | 0.2874 | 0.2986 | 0.3071 |
| $m_{clust} = 0.8$ | **<u>0.2575</u>** | 0.2800 | 0.2846 | 0.2950 | 0.3015 |
| $m_{clust} = 1$ | 0.2583 | **0.2773** | **0.2844** | **0.2911** | **0.2985** |

(b) Salinas

**Table 2**
Evaluation of superpixel segmentation on (a) SalinasA and (b) Salinas datasets with under-segmentation error. The table shows the error considering difference values of $m$ and $m_{clust}$. For each column, cells in orange show the maximum values (worst results). Values in bold-face represent the best value for each column while underlined values show the best value overall.

| Dataset | Spectral original | Superpixels original | Spectral PCA | Superpixels PCA |
|---|---|---|---|---|
| Pavia Center | 102 | 104 | 15 | 7 |
| Pavia University | 103 | 105 | 16 | 8 |
| Salinas | 204 | 206 | 13 | 9 |
| SalinasA | 204 | 206 | 17 | 10 |

**Table 3**
The number of bands before and after the application of PCA is shown for both the original spectral information of the image and the information of the superpixels centers.

| Clustering method | Input Information | PCA | ARI | NMI | F1 |
|---|---|---|---|---|---|
| K-means | Spectral | No | 0.78 | 0.76 | 0.82 |
| K-means | Spectral | Yes | 0.78 | 0.76 | 0.82 |
| K-means | Superpixels | No | 0.78 | 0.70 | 0.80 |
| K-means | Superpixels | Yes | 0.77 | 0.69 | 0.81 |
| K-means | Spectral + Superpixels | No | 0.78 | 0.74 | 0.81 |
| K-means | Spectral + Superpixels | Yes | 0.79 | 0.77 | 0.84 |
| Mean-shift | Spectral | No | 0.80 | 0.77 | 0.85 |
| Mean-shift | Spectral | Yes | 0.80 | 0.78 | 0.84 |
| Mean-shift | Superpixels | No | 0.82 | 0.75 | 0.88 |
| Mean-shift | Superpixels | Yes | 0.82 | 0.75 | 0.88 |
| Mean-shift | Spectral + Superpixels | No | **0.88** | **0.87** | **0.90** |
| Mean-shift | Spectral + Superpixels | Yes | **0.88** | 0.86 | 0.89 |

**Table 4**
The table shows the performance of our technique on Pavia Center dataset using different combinations of methods and inputs. In the mean-shift methods, the table reports the results on each metric with a fixed bandwidth, that has been selected considering the best results achieved on NMI.

| Clustering method | Input Information | PCA | ARI | NMI | F1 |
|---|---|---|---|---|---|
| K-means | Spectral | No | 0.31 | 0.57 | 0.60 |
| K-means | Spectral | Yes | 0.32 | 0.58 | 0.65 |
| K-means | Superpixels | No | 0.32 | 0.54 | 0.63 |
| K-means | Superpixels | Yes | 0.32 | 0.52 | 0.63 |
| K-means | Spectral + Superpixels | No | 0.33 | 0.62 | 0.65 |
| K-means | Spectral + Superpixels | Yes | 0.33 | 0.56 | 0.62 |
| Mean-shift | Spectral | No | 0.47 | 0.58 | 0.74 |
| Mean-shift | Spectral | Yes | 0.47 | 0.58 | 0.74 |
| Mean-shift | Superpixels | No | 0.42 | 0.49 | 0.74 |
| Mean-shift | Superpixels | Yes | 0.38 | 0.44 | 0.71 |
| Mean-shift | Spectral + Superpixels | No | **0.59** | **0.72** | **0.84** |
| Mean-shift | Spectral + Superpixels | Yes | 0.57 | 0.69 | 0.82 |

**Table 5**
The table shows the performance of our technique on Pavia University dataset using different combinations of methods and inputs. In the mean-shift methods, the table reports the results on each metric with a fixed bandwidth, that has been selected considering the best results achieved on NMI.

| Clustering method | Input Information | PCA | ARI | NMI | F1 |
|---|---|---|---|---|---|
| K-means | Spectral | No | 0.56 | 0.80 | 0.74 |
| K-means | Spectral | Yes | 0.59 | 0.81 | 0.77 |
| K-means | Superpixels | No | 0.60 | 0.81 | 0.77 |
| K-means | Superpixels | Yes | 0.66 | 0.84 | 0.81 |
| K-means | Spectral + Superpixels | No | 0.62 | 0.82 | 0.79 |
| K-means | Spectral + Superpixels | Yes | 0.56 | 0.82 | 0.76 |
| Mean-shift | Spectral | No | 0.69 | 0.88 | 0.84 |
| Mean-shift | Spectral | Yes | 0.69 | 0.87 | 0.84 |
| Mean-shift | Superpixels | No | 0.70 | 0.86 | 0.84 |
| Mean-shift | Superpixels | Yes | 0.71 | 0.87 | 0.84 |
| Mean-shift | Spectral + Superpixels | No | **0.85** | **0.91** | **0.90** |
| Mean-shift | Spectral + Superpixels | Yes | 0.82 | **0.91** | 0.89 |

**Table 6**
The table shows the performance of our technique on Salinas dataset using different combinations of methods and inputs. In the mean-shift methods, the table reports the results on each metric with a fixed bandwidth, that has been selected considering the best results achieved on NMI.

| Clustering method | Input Information | PCA | ARI | NMI | F1 |
|---|---|---|---|---|---|
| K-means | Spectral | No | 0.80 | 0.90 | 0.89 |
| K-means | Spectral | Yes | 0.80 | 0.90 | 0.89 |
| K-means | Superpixels | No | 0.82 | 0.90 | 0.90 |
| K-means | Superpixels | Yes | 0.58 | 0.79 | 0.77 |
| K-means | Spectral + Superpixels | No | 0.82 | 0.90 | 0.90 |
| K-means | Spectral + Superpixels | Yes | 0.82 | 0.90 | 0.90 |
| Mean-shift | Spectral | No | 0.73 | 0.84 | 0.83 |
| Mean-shift | Spectral | Yes | 0.73 | 0.84 | 0.83 |
| Mean-shift | Superpixels | No | 0.69 | 0.80 | 0.83 |
| Mean-shift | Superpixels | Yes | 0.72 | 0.83 | 0.85 |
| Mean-shift | Spectral + Superpixels | No | **0.90** | **0.95** | **0.95** |
| Mean-shift | Spectral + Superpixels | Yes | **0.90** | **0.95** | **0.95** |

**Table 7**
The table shows the performance of our technique on SalinasA dataset using different combinations of methods and inputs. In the mean-shift methods, the table reports the results on each metric with a fixed bandwidth, that has been selected considering the best results achieved on NMI.

| | Pavia Center ARI - NMI | Pavia Univ. ARI - NMI | Salinas ARI - NMI | SalinasA ARI - NMI | **Average** ARI - NMI | Require number of classes |
|---|---|---|---|---|---|---|
| K-means (Obeid et al. (2021)) | 0.85 - 0.82 | 0.40 - 0.63 | 0.63 - 0.83 | 0.67 - 0.78 | 0.64 - 0.77 | Yes |
| GMM (Obeid et al. (2021)) | 0.77 - 0.74 | 0.29 - 0.53 | 0.53 - 0.79 | 0.78 - 0.87 | 0.59 - 0.73 | Yes |
| HNMF (Gillis et al. (2014)) | 0.85 - 0.77 | 0.38 - 0.57 | 0.53 - 0.79 | 0.78 - 0.87 | 0.64 - 0.75 | Yes |
| SMCE (Elhamifar and Vidal (2013)) | 0.80 - 0.77 | 0.31 - 0.56 | 0.57 - 0.78 | 0.76 - 0.81 | 0.61 - 0.73 | Yes |
| DLSS (Murphy and Maggioni (2018)) | 0.52 - 0.42 | 0.49 - 0.57 | 0.37 - 0.39 | 0.63 - 0.81 | 0.55 - 0.50 | Yes |
| 3D-CAE (Nalepa et al. (2020)) | 0.96 - 0.86 | 0.36 - 0.59 | 0.67 - 0.85 | 0.77 - 0.87 | 0.69 - 0.79 | Yes |
| DEC (Xie et al. (2016)) | 0.83 - 0.80 | 0.41 - 0.67 | 0.57 - 0.80 | 0.78 - 0.87 | 0.65 - 0.79 | Yes |
| BDEC (Obeid et al. (2021)) | **0.97** - **0.91** | **0.60** - 0.70 | 0.68 - 0.87 | 0.81 - 0.87 | 0.77 - 0.84 | Yes |
| **OUR_BW** | 0.81 - 0.80 | 0.53 - 0.70 | 0.67 - 0.87 | 0.82 - 0.92 | 0.71 - 0.82 | **No** |
| **OUR** | 0.88 - 0.87 | 0.59 - **0.72** | **0.85** - **0.91** | **0.90** - **0.95** | **0.81** - **0.86** | **No** |

**Table 8**
Comparison of our method with other methods in the state of art using NMI and ARI. The table shows the ARI and NMI values achieved by every method on the datasets of Pavia Center, Pavia University, Salinas, and SalinasA.

| Variance | Pavia Center ARI - NMI | Pavia Univ. ARI - NMI | Salinas ARI - NMI | SalinasA ARI - NMI | **Average** ARI - NMI |
|---|---|---|---|---|---|
| 0 | 0.81 - 0.80 | 0.53 - 0.70 | 0.67 - 0.87 | 0.82 - 0.92 | 0.71 - 0.82 |
| 0.01 | 0.81 - 0.78 | 0.53 - 0.69 | 0.54 - 0.79 | 0.90 - 0.95 | 0.70 - 0.80 |
| 0.05 | 0.81 - 0.78 | 0.56 - 0.68 | 0.48 - 0.74 | 0.82 - 0.90 | 0.67 - 0.78 |
| 0.1 | 0.84 - 0.81 | 0.56 - 0.65 | 0.46 - 0.71 | 0.73 - 0.84 | 0.65 - 0.75 |
| 0.25 | 0.84 - 0.81 | 0.55 - 0.65 | 0.45 - 0.71 | 0.34 - 0.56 | 0.55 - 0.68 |
| 0.5 | 0.83 - 0.80 | 0.54 - 0.64 | 0.38 - 0.62 | 0.24 - 0.51 | 0.50 - 0.64 |

**Table 9**
The table shows the robustness of our method to the addition of Gaussian Noise for each of the datasets with different variances.

| Pixel density | Pavia Center ARI - NMI | Pavia Univ. ARI - NMI | Salinas ARI - NMI | SalinasA ARI - NMI | **Average** ARI - NMI |
|---|---|---|---|---|---|
| 0 | 0.81 - 0.80 | 0.53 - 0.70 | 0.67 - 0.87 | 0.82 - 0.92 | 0.71 - 0.82 |
| 0.01 | 0.81 - 0.79 | 0.53 - 0.70 | 0.66 - 0.87 | 0.75 - 0.90 | 0.69 - 0.82 |
| 0.05 | 0.81 - 0.80 | 0.52 - 0.68 | 0.66 - 0.86 | 0.82 - 0.92 | 0.70 - 0.82 |
| 0.1 | 0.80 - 0.78 | 0.52 - 0.68 | 0.66 - 0.86 | 0.90 - 0.95 | 0.72 - 0.82 |
| 0.25 | 0.81 - 0.78 | 0.50 - 0.67 | 0.58 - 0.82 | 0.90 - 0.95 | 0.70 - 0.81 |
| 0.5 | 0.80 - 0.78 | 0.57 - 0.69 | 0.52 - 0.77 | 0.83 - 0.90 | 0.66 - 0.79 |

**Table 10**
The table shows the robustness of our method to the addition of Impulsive Noise for each of the datasets with different densities of pixels.

| Poisson noise | Pavia Center<br>ARI - NMI | Pavia Univ.<br>ARI - NMI | Salinas<br>ARI - NMI | SalinasA<br>ARI - NMI | **Average**<br>ARI - NMI |
|---|---|---|---|---|---|
| No | 0.81 - 0.80 | 0.53 - 0.70 | 0.67 - 0.87 | 0.82 - 0.92 | 0.71 - 0.82 |
| Yes | 0.81 - 0.80 | 0.52 - 0.69 | 0.66 - 0.87 | 0.80 - 0.92 | 0.70 - 0.82 |

**Table 11**
The table shows the robustness of our method to the addition of Poisson Noise for each of the datasets.

| | Pavia Center | Pavia Univ. | Salinas | SalinasA | **Average** |
|---|---|---|---|---|---|
| Augmented H-SLIC | 1097.01s | 343.16s | 388.67s | 43.44s | 468.07s |
| Unsupervised Segmentation | 835.57s | 218.53s | 303.06s | 19.04s | 344.05s |
| **Total** | 1944.61s | 564.87s | 693.88s | 62.94s | 816.58s |

**Table 12**
The table shows the time computation in seconds for Pavia Center, Pavia University, Salinas, and SalinasA datasets, for each step and the entire pipeline.